\documentclass{article}

\PassOptionsToPackage{numbers,square,sort&compress}{natbib}
\usepackage[main,final]{neurips_2026}
\usepackage[utf8]{inputenc}
\usepackage[T1]{fontenc}
\usepackage{url}
\usepackage{booktabs}
\usepackage{amsfonts}
\usepackage{amsmath}
\usepackage{amssymb}
\usepackage{amsthm}
\usepackage{nicefrac}
\usepackage{microtype}
\usepackage{xcolor}
\usepackage{graphicx}
\usepackage{wrapfig}
\usepackage{multirow}
\usepackage{array}
\usepackage{tabularx}
\usepackage{enumitem}
\usepackage{algorithm}
\usepackage{algorithmic}
\usepackage{placeins}
\usepackage{tikz}
\usepackage[hidelinks]{hyperref}

\newcommand{\modelname}{BusemannFormer}

\newcommand{\modelsym}{\modelname{}-S}
\newcommand{\modelld}{\modelname{}-LD}

\newtheorem{proposition}{Proposition}

\title{\modelname: Horospherical Self-Attention for Hyperbolic Graph Transformers}

\author{
  Youheng Yao$^{1}$\thanks{equal contribution} \quad Ziyao Zeng$^{1}$\footnotemark[1] \quad
  Wenbo Liao$^{2}$ \quad Tianqi Wang$^{1}$\\
  $^{1}$Yale University \quad $^{2}$The Chinese University of Hong Kong\\
  \texttt{youheng.yao@yale.edu} \quad \texttt{ziyao.zeng@yale.edu}\\
  \texttt{liao.wenbo98@gmail.com} \quad \texttt{tq.wang@yale.edu}
}

\begin{document}
\maketitle

\begin{abstract}
Hyperbolic neural networks excel on hierarchical graph data, yet existing
hyperbolic Transformers largely adapt conventional attention without
explicitly exploiting the ideal boundary, a geometric structure closely
tied to tree hierarchy. We propose \textbf{Horospherical Self-Attention}
(HSA): each query selects an ideal-boundary direction and scores keys by
their Busemann depth along that direction, yielding a token-to-token
hyperbolic attention kernel. We establish a flat-limit connection to
dot-product attention, characterize score level sets as horospheres,
and bound key-side gradients. HSA is integrated into
\textbf{BusemannFormer}, a fully hyperbolic graph Transformer, with a
log-depth variant that uses query depth to modulate attention
concentration. We further introduce a finite-boundary-dictionary
extension supporting linear-complexity global attention.
Experiments on hierarchical and citation graphs show that \modelld{}
outperforms the evaluated baselines on several node-classification
benchmarks and achieves competitive performance across diverse graph
structures.
\end{abstract}

\section{Introduction}
\label{sec:intro}

Many graph datasets exhibit hierarchical or tree-like structure, including
taxonomies~\citep{nickel2017poincare}, disease-propagation trees and certain
citation networks~\citep{chami2019hgcn}, and hierarchical relations in
knowledge graphs~\citep{balazevic2019multi}.
Hyperbolic space is the natural candidate for representing such data, in the
sense of Gromov hyperbolicity~\citep{bridson1999metric}.
Hyperbolic graph neural networks have demonstrated strong performance on
node classification and link prediction tasks~\citep{chami2019hgcn}, and
Hypformer extends Transformer architectures fully into hyperbolic
space~\citep{yang2024hypformer}.

\begin{figure}[t]
  \centering
  \includegraphics[width=\linewidth]{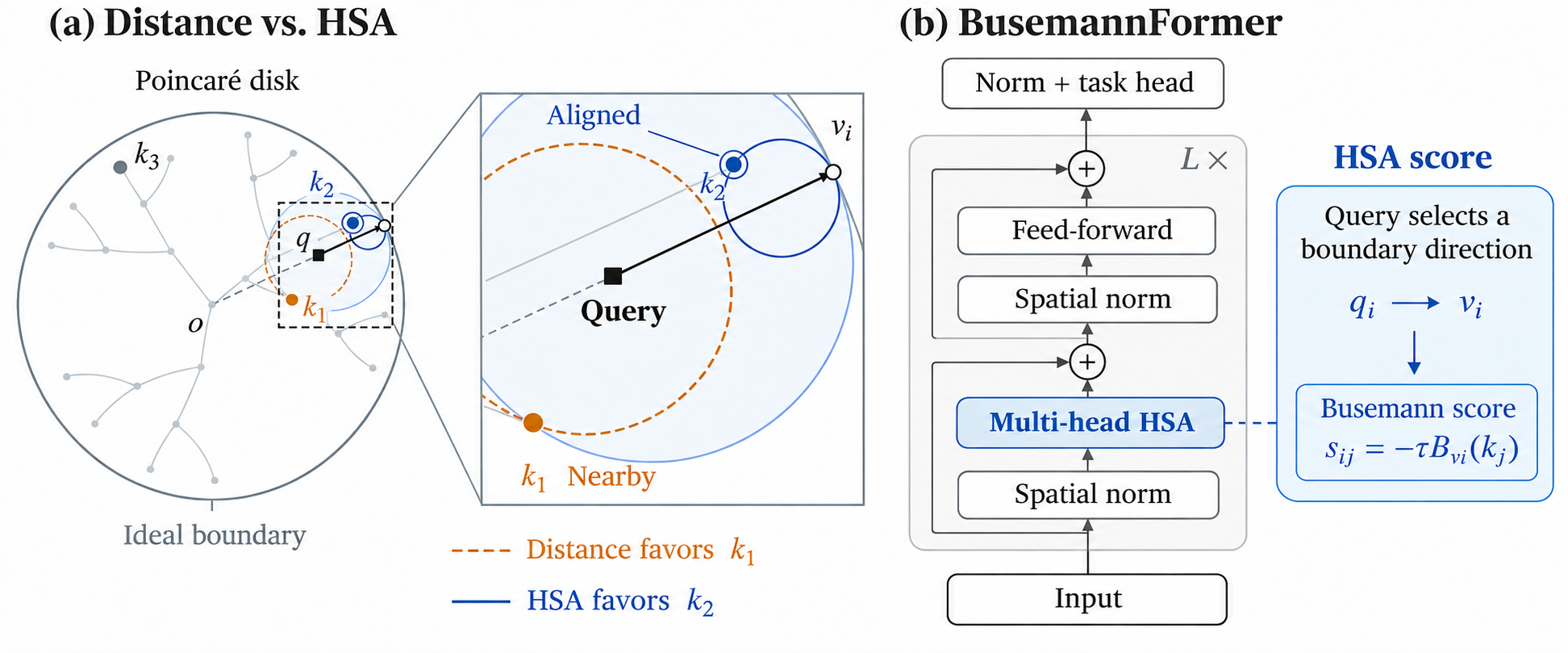}
  \caption{\textbf{BusemannFormer overview.}
  (a) In the Poincar\'e disk $\mathbb{H}^2$, HSA favors directionally
  aligned $k_2$ over nearby $k_1$. Orange/blue curves mark equal
  hyperbolic distance/Busemann depth.
  (b) The hyperbolic Transformer uses HSA: each query selects a boundary
  direction and scores keys by Busemann depth; $\oplus$ denotes spatial
  addition and lift.}
  \label{fig:hsa-motivation-teaser}
\end{figure}

However, representing a hierarchy and retrieving information along it are
different design choices. A hyperbolic representation alone does not
specify how a query should select information according to depth along a
particular direction. This motivates an attention mechanism that makes
directional depth explicit. Figure~\ref{fig:hsa-motivation-teaser}
illustrates this motivation and the model overview.

The ideal boundary of hyperbolic space provides a geometric construction
for this purpose. Each boundary point determines a Busemann function:
a signed depth coordinate whose level sets are horospheres centered at the boundary point. We propose
\textbf{Horospherical Self-Attention (HSA)}, in which each query selects a
boundary direction and scores keys by their Busemann depth along that
direction. Intuitively, a query asks: \emph{which keys lie farther along my
selected branch?} This is a hyperbolic analogue of Euclidean projection:
just as a Euclidean query selects direction $v = q/\|q\|$ and ranks
keys by the dot product $\langle v, k\rangle$, an HSA query selects a boundary direction and
ranks keys by horospherical depth.
Dot-product attention is recovered when $K\to 0^-$
(Proposition~\ref{prop:euclidean-limit}).

Busemann functions have been used in deep learning for dimensionality
reduction~\citep{chami2021horopca}, prototype
classification~\citep{ghadimi2021busemann}, decision
boundaries~\citep{fan2023horosphere}, and fixed-direction layers~\citep{chen2026hbnn}.
All of these use fixed or class-level boundary directions.
None treats the Busemann function as a \emph{token-to-token self-attention
kernel} where each query token chooses its own boundary direction at
inference time.

We address this gap with HSA and integrate it into
\textbf{\modelname{}}, a hyperbolic graph Transformer.
Code is available at \url{https://github.com/LionelYao/Busemannformer}.

Experiments on hierarchical and citation graphs show that
\modelld{} achieves the highest mean among the evaluated
baselines on four of five node-classification benchmarks,
including \textbf{92.66\%} macro-F1 on Disease
(Table~\ref{tab:nc}). Beyond these benchmarks, Busemann modulation of Hypformer's
native attention reaches \textbf{72.912\%} test accuracy on
ogbn-arxiv (Table~\ref{tab:linear-performance}).
Complementary efficiency experiments quantify the runtime and
memory savings of boundary-dictionary attention. 

\paragraph{Contributions.}
\begin{enumerate}[itemsep=2pt, leftmargin=*]
  \item \textbf{Boundary-directed attention.}
    We introduce HSA and integrate it into a hyperbolic graph Transformer,
    providing an explicit geometric interpretation of query-dependent
    retrieval.
  \item \textbf{\modelname{}.} A full architecture with HypLinear projections,
        an optional parallel Lorentz GNN branch, and a BMLR classifier
        (Section~\ref{sec:arch}).
  \item \textbf{Theory.} HSA recovers dot-product attention as $K\to0^-$;
        score level sets are horospheres; key-side gradients are bounded
        (Section~\ref{sec:theory}).
  \item \textbf{Experiments.} Benchmarks on five datasets with controlled
        score and architecture ablations (Section~\ref{sec:experiments}).
\end{enumerate}

\section{Background}
\label{sec:background}

We work in the Lorentz model of hyperbolic space with curvature
\(K=-c^2\), where \(c>0\). Set
\[\langle x,y\rangle_{\mathbb L}
  =-x_0y_0+\langle x_s,y_s\rangle_{\mathbb R^d} \quad \text{and} \quad
  \mathbb L_K^d
  =
  \{x=(x_0,x_s)\in\mathbb R^{d+1}:
  \langle x,x\rangle_{\mathbb L}=-c^{-2},\ x_0>0\}.
\]
After fixing the basepoint \(o_K=(c^{-1},0)\), we identify the ideal boundary
\(\partial_\infty\mathbb L_K^d\) with \(\mathbb S^{d-1}\).  For
\(\xi\in\mathbb S^{d-1}\), the normalized Busemann function is
\begin{equation}
  B_\xi^{(K)}(x)
  =
  \frac1c\log\!\left(c\bigl(x_0-\langle x_s,\xi\rangle\bigr)\right),
  \qquad x\in\mathbb L_K^d.
  \label{eq:busemann}
\end{equation}
Its level sets \(H_{\xi,\alpha}^{(K)}=\{x:B_\xi^{(K)}(x)=\alpha\}\) are
horospheres centred at \(\xi\), and along the unit-speed geodesic ray toward
\(\xi\) one has \(B_\xi^{(K)}(\gamma_\xi(t))=-t\).  Thus
\(-B_\xi^{(K)}\) is a hyperbolic directional-depth coordinate.  When the
curvature is clear, we write \(B_\xi\) and \(\mathbb L^d\) for brevity.  The
hyperbolic distance, boundary parametrisation, exponential and logarithmic
maps, centroid projection, horosphere equations, and the precise
normal-coordinate flat limit are collected in Appendix~\ref{app:hyperbolic}.

\paragraph{Hyperbolic linear and Busemann layers.}
HGCN~\citep{chami2019hgcn} introduces \textbf{HypLinear}, which applies a
Euclidean linear map to the spatial part of a hyperboloid point and reprojects
to $\mathbb L^d$. HBNN~\citep{chen2026hbnn} introduces the related
\textbf{Busemann fully connected} (BFC) layer,
\begin{equation}
  r_m(x)=-\alpha_m B_{\xi_m}^{(K)}(x)+b_m,
  \qquad \xi_m\in\mathbb S^{d-1},\quad \alpha_m>0,
\end{equation}
and the \textbf{BMLR} classifier, whose logits have horosphere decision
boundaries. In \modelname{}, we use \textbf{HypLinear} for the Q, K, V, output,
and FFN projections, and \textbf{BMLR} as the final node classifier.
\section{Horospherical Self-Attention}
\label{sec:method}

HSA uses Busemann functions to define query-dependent token-to-token
scores: each query selects an ideal-boundary direction, and each key
is scored by its depth toward that boundary point. This makes branch
direction and hierarchical depth explicit in attention. We first define
the directed score and its log-depth and bidirectional variants, then
describe value aggregation and multi-head attention.

\paragraph{Setup.}
Given token representations
$x_1,\ldots,x_N\in\mathbb{L}_{K}^{d}$ with curvature $K=-c^2$,
each head $h$ produces hyperbolic queries, keys, and values:
\begin{equation}
\begin{aligned}
q_i^{(h)} &= \operatorname{HypLinear}_{Q}^{(h)}(x_i),\\
k_i^{(h)} &= \operatorname{HypLinear}_{K}^{(h)}(x_i),\\
z_i^{(h)} &= \operatorname{HypLinear}_{V}^{(h)}(x_i),
\end{aligned}
\qquad
q_i^{(h)},k_i^{(h)},z_i^{(h)}
\in\mathbb{L}_{K}^{d_h},
\end{equation}
where $d_h=d/H$ and $H$ is the number of heads.
We suppress the head index when the meaning is clear.

\subsection{Query-Dependent Boundary Direction}

Each query token $q_i^{(h)} = (q_{i,0}^{(h)}, q_{i,s}^{(h)})$ with nonzero spatial component induces an
ideal-boundary direction by normalising its spatial part:
\begin{equation}
v_i^{(h)}
=
\frac{q_{i,s}^{(h)}}{\|q_{i,s}^{(h)}\|_2}
\in S^{d_h-1}.
\label{eq:boundary-dir}
\end{equation}
For numerical stability, the implementation uses
\[
  v_i^{(h)} = \frac{q_{i,s}^{(h)}}{\max\{\|q_{i,s}^{(h)}\|,\epsilon\}}, \qquad \epsilon = 10^{-12}
\]
Geometrically, $v_i^{(h)}$ is a \emph{direction} on the ideal boundary,
selected dynamically by query token $i$.

\subsection{Horospherical Attention Score}

The directed attention score between query $i$ and key $j$
in head $h$ is
\begin{equation}
s_{ij}^{(h)}
=
-\tau_h B_{v_i^{(h)}}\!\left(k_j^{(h)}\right)
=
-\frac{\tau_h}{c}
\log\!\left[
c\left(k_{j,0}^{(h)}
-\left\langle k_{j,s}^{(h)},v_i^{(h)}\right\rangle\right)
\right],
\label{eq:hsa-score}
\end{equation}
where $\tau_h>0$ is a learnable head-specific temperature,
initialized to $\tau_h=\sqrt{d_h}$.
The attention weight is $a_{ij}^{(h)} = \operatorname{softmax}_j(s_{ij}^{(h)})$.

For a fixed query, the score level sets are horospheres
centered at its selected ideal-boundary point. Along the
unit-speed geodesic ray $\gamma_{v_i^{(h)}}(t)$ from the
basepoint, $B_{v_i^{(h)}}\!\left(\gamma_{v_i^{(h)}}(t)\right)=-t$,
so the attention score increases as $\tau_h t$. The query specifies
the direction, and keys are ranked by their Busemann depth along it.

\subsection{Log-Depth Horospherical Attention (\modelld{})}
\label{sec:ld-score}

The directed score~\eqref{eq:hsa-score} uses the same temperature
for every query within a head. At unit curvature,
$q_{i,0}=\cosh r_i$, where $r_i$ is the query's distance from
the basepoint. We use this radial information to define a
query-dependent inverse temperature:
\begin{equation}
s_{ij}^{(h),\mathrm{ld}}
=
-\tau_h
\Bigl(1+\gamma_h
\log\max\{q_{i,0}^{(h)},1+\epsilon\}\Bigr)
B_{v_i^{(h)}}\!\bigl(k_j^{(h)}\bigr),
\label{eq:hsa-ld}
\end{equation}
where $\epsilon=10^{-6}$ sets the lower clipping threshold and
$\gamma_h=\operatorname{softplus}(\widehat\gamma_h)$ is a
learnable per-head depth gain, initialized with
$\widehat\gamma_h=-3$ ($\gamma_h\approx0.049$).
Thus \modelld{} learns a query-dependent sharpening of attention.
The multiplicative factor preserves the ordering of keys along the selected boundary direction,
while increasing the score scale for deeper queries, whose boundary directions
are more stable.

\subsection{Bidirectional Busemann Score}
\label{sec:sym-score}

A bidirectional variant incorporates both the depth of the key
along the query direction and the depth of the query along the
key direction:
\begin{equation}
\begin{aligned}
s_{ij}^{(h),\mathrm{bi}}
=-\frac{1}{2}\Big[
a_h(q_i^{(h)})
B_{\xi(q_i^{(h)})}\!\left(k_j^{(h)}\right)
+{}&
a_h(k_j^{(h)})
B_{\xi(k_j^{(h)})}\!\left(q_i^{(h)}\right)
\Big],
\end{aligned}
\label{eq:hsa-sym}
\end{equation}
where $a_h(x)=\tau_h$ for the direct variant and
$a_h(x)=\tau_h g_h(x)$ for the log-depth variant.

\subsection{Value Aggregation}
\label{sec:aggregation}

We aggregate full Lorentz value vectors using the Lorentz
centroid~\citep{lou2020frechet}:
\begin{align}
m_i^{(h)}
&=
\sum_{j\in\mathcal{S}_i}a_{ij}^{(h)}z_j^{(h)},
\qquad
o_i^{(h)}
=
\Pi_c(m_i^{(h)}),
\label{eq:lorentz-centroid}\\
\Pi_c(m)
&=
\frac{m}
{c\sqrt{-\langle m,m\rangle_{\mathbb L}}}.
\end{align}
Nonnegative weights with positive total mass make $m$ future
timelike, so $\Pi_c(m)$ lies on the hyperboloid.
This efficient extrinsic aggregation is not generally the
intrinsic Fr\'echet mean.

\subsection{Busemann Multi-Head Attention (BMHA)}

We concatenate the spatial components of the head outputs and
reconstruct the time coordinate before the output projection:
\begin{align}
u_i
&=
\operatorname{Concat}
\left(o_{i,s}^{(1)},\ldots,o_{i,s}^{(H)}\right),
\\
\operatorname{BMHA}(X)_i
&=
\operatorname{HypLinear}_{\mathrm{out}}
\left(\operatorname{Lift}_c(u_i)\right),
\label{eq:bmha-output}\\
\operatorname{Lift}_c(s)
&=
\left(\sqrt{c^{-2}+\|s\|_2^2},\,s\right).
\label{eq:spatial-lift}
\end{align}
\section{\modelname{} Architecture}
\label{sec:arch}

\modelname{} integrates HSA into a hyperbolic graph
Transformer. Node features are mapped to spatial coordinates and
lifted to the hyperboloid:
\begin{equation}
x_i^{(0)}
=
\operatorname{Lift}_{c_{\mathrm{in}}}
\left(W_{\mathrm{in}}h_i+b_{\mathrm{in}}\right).
\end{equation}
A stack of $L$ blocks combines attention, an optional graph
convolution branch, and a feed-forward network. Token
representations are returned to the appropriate hyperboloid
after each sublayer.

\paragraph{Manifold-preserving modules.}
Spatial affine transformations, normalization, and residuals are
\begin{align}
\operatorname{HypLinear}_c(x)
&=
\operatorname{Lift}_c(Wx_s+b),\\
\operatorname{SpatialLN}_c(x)
&=
\operatorname{Lift}_c\!\left(\operatorname{LN}(x_s)\right),\\
\operatorname{Res}_c(x,y)
&=
\operatorname{Lift}_c(x_s+y_s).
\end{align}
The feed-forward network is
\begin{equation}
\operatorname{FFN}_c(x)
=
\operatorname{HypLinear}_{2,c}
\left(
\sigma_c\!\left(\operatorname{HypLinear}_{1,c}(x)\right)
\right),
\end{equation}
where
$\sigma_c(x)=\operatorname{Lift}_c(\operatorname{GELU}(x_s))$.
Its hidden spatial width is $4d$.
These coordinate-based operations preserve the hyperboloid
constraint without requiring tangent-space logarithmic and
exponential maps.

\paragraph{Lorentz GNN branch.}
When enabled, a parallel one-hop graph-convolution branch uses
the same pre-normalized input $\widehat X$ as attention.
Let $\widehat A$ be the row-normalized adjacency with self-loops.
The branch computes
\begin{equation}
\operatorname{LorentzGNN}_c(\widehat X,\widehat A)_i
=
\operatorname{HypLinear}_c
\left(
\Pi_c\!\left(\sum_j\widehat A_{ij}\widehat x_j\right)
\right).
\label{eq:lorentz-gnn}
\end{equation}
The result is incorporated through a spatial residual.

\paragraph{Curvature and block structure.}
When curvature is learned, block $\ell$ stores a log curvature
magnitude $\theta_\ell$:
\begin{equation}
\rho_\ell=e^{\theta_\ell},
\qquad
c_\ell=\sqrt{\rho_\ell},
\qquad
K_\ell=-c_\ell^2=-\rho_\ell.
\end{equation}
All submodules in the block use $c_\ell$.
Fixed-curvature experiments freeze these parameters.
Time-coordinate reconstruction accommodates changes in curvature
between blocks while retaining the spatial coordinates.

Omitting dropout for clarity, one block is
\begin{align}
\widehat X
&=
\operatorname{SpatialLN}_{c_\ell}^{(1)}
\left(X^{(\ell)}\right),
\nonumber\\
U
&=
\operatorname{Res}_{c_\ell}
\left(
X^{(\ell)},\mathcal T_{c_\ell}(\widehat X)
\right),
\nonumber\\
U
&\leftarrow
\operatorname{Res}_{c_\ell}
\left(
U,\operatorname{LorentzGNN}_{c_\ell}(\widehat X,\widehat A)
\right)
\quad\text{if enabled},
\nonumber\\
X^{(\ell+1)}
&=
\operatorname{Res}_{c_\ell}
\left(
U,
\operatorname{FFN}_{c_\ell}
\left(\operatorname{SpatialLN}_{c_\ell}^{(2)}(U)\right)
\right).
\label{eq:block}
\end{align}
For the original model, $\mathcal T=\operatorname{BMHA}$.
Attention and graph convolution consume the same $\widehat X$;
the feed-forward network consumes the updated residual state.

\paragraph{Graph support and prediction.}
For graph-local attention, $\mathcal S_i$ contains node $i$ and
nodes within a specified number of graph hops. Unrestricted
attention uses all nodes as keys. The support is an experimental
choice rather than a change to the HSA score.

After the final block, spatial layer normalization is followed
by a per-node BMLR-style classifier. The implemented class logits are
\begin{equation}
\ell_t(x)
=
-\alpha_t
\log\max\{x_0-x_s^\top\nu_t,10^{-6}\}+b_t,
\qquad
\nu_t=\frac{u_t}{\|u_t\|_2}.
\end{equation}
The class scales $\alpha_t$ are learned directly and may be signed;
we distinguish this readout from a strictly positive-scale BMLR
parameterization.

\subsection{Linear Boundary Attention}
\label{sec:linear-boundary-attention}

To mitigate the quadratic cost of the attention score,
we introduce a finite dictionary of learned boundary directions.
Each head learns $M$ directions
$\xi_m=r_m/\|r_m\|_2$, with raw vectors $r_m$ randomly initialized
and optimized end-to-end.

\paragraph{Boundary features.}
We describe the single-scale, unit-curvature configuration, suppressing
head indices. Each key is encoded by exponentiated Busemann depths:
\begin{equation}
\psi_m(k)
=
e^{-aB_{\xi_m}(k)}
=
\left(k_0-k_s^\top\xi_m\right)^{-a},
\label{eq:boundary-key-features}
\end{equation}
where the query-independent exponent $a>0$ is shared across dictionary
directions. Queries assign nonnegative routing weights to these directions:
\begin{equation}
\phi_m(q)
=
\operatorname{softmax}_m
\left(\beta g(q)\,\xi(q)^\top\xi_m+b_m\right),
\label{eq:boundary-query-routing}
\end{equation}
where $\beta>0$ and $b_m$ are learned parameters, and
\[
g(q)=1+\gamma\log\max\{q_0,1+\epsilon\},
\qquad \epsilon=10^{-6}.
\]
Together, these maps define the finite separable kernel
\begin{equation}
K_M(q,k)
=
\phi(q)^\top\psi(k)
=
\sum_{m=1}^{M}\phi_m(q)\psi_m(k).
\label{eq:boundary-dictionary-kernel}
\end{equation}

\paragraph{Associative linear attention.}
For hyperbolic values $z_j$, we accumulate the shared statistics
\begin{equation}
S=\sum_{j=1}^{N}\psi(k_j)z_j^\top,
\qquad
b=\sum_{j=1}^{N}\psi(k_j).
\end{equation}
The normalized ambient aggregate and hyperbolic output are
\begin{equation}
y_i
=
\frac{S^\top\phi(q_i)}{b^\top\phi(q_i)},
\qquad
o_i=\Pi_1(y_i).
\label{eq:boundary-linear-aggregation}
\end{equation}
This computes normalized attention for $K_M$ exactly without
constructing an $N\times N$ matrix.
Spatial head outputs are concatenated, lifted, and projected
as in BMHA. Across $H$ heads, encoding and aggregation cost
$O(HNMd_h)$, excluding projections, and are linear in $N$
for fixed dictionary size and width.

\subsection{Graph Instantiation}
\label{sec:graph-instantiation}

For graph data, connectivity enters through a graph-local
attention support and an optional Lorentz graph-convolution
branch. The scalable variant additionally incorporates global
boundary-dictionary attention.

\paragraph{Graph-supported HSA.}
Let $\mathcal S_i$ contain node $i$ and nodes within a specified
number of graph hops. Local attention uses the HSA scores from
Section~\ref{sec:method}, normalized over this support:
\begin{equation}
a_{ij}^{(h),\mathrm{local}}
=
\frac{
\mathbf{1}_{\{j\in\mathcal S_i\}}\exp(s_{ij}^{(h)})
}{
\sum_{r\in\mathcal S_i}\exp(s_{ir}^{(h)})
}.
\label{eq:graph-local-hsa}
\end{equation}
We evaluate only the
$E_{\mathcal S}=\sum_i|\mathcal S_i|$ supported pairs, apply
query-wise segment softmax, and aggregate their values.
This computes the same operator as dense masked attention in
exact arithmetic, with cost $O(H E_{\mathcal S}d_h)$,
excluding projections and support construction.

\paragraph{Optional Lorentz GNN.}
The optional Lorentz GNN branch is defined in~\eqref{eq:lorentz-gnn}.
Its output is added through a spatial residual after the attention
residual and before the feed-forward network. This branch is distinct
from graph-local attention and can be disabled independently.

\section{Theoretical Analysis}
\label{sec:theory}

We establish two structural properties of HSA. Throughout this section,
the curvature is \(K=-c^2\) with
\(c>0\), and an HSA head uses the score
\[
  s_{ij}^{(h)}=-\tau_h B_{v_i^{(h)}}^{(K)}(k_j^{(h)}).
\]
Detailed proofs and auxiliary facts---horosphere level sets, gradient
stability, geodesic depth ordering, branch separation, and the precise
symmetry condition for \modelsym{}---are collected in Appendix~\ref{app:proofs}.
The background formulas for the Lorentz model are in Appendix~\ref{app:hyperbolic}.

\begin{proposition}[Flat limit in normal coordinates]
\label{prop:euclidean-limit}
Let \(\iota_c(y)=\operatorname{Exp}_{o_K}^{K}(0,y)\) be the normal-coordinate
embedding of \(y\in\mathbb R^{d_h}\) into \(\mathbb L_K^{d_h}\).  For every
\(\xi\in\mathbb S^{d_h-1}\), uniformly for \(y\) in compact subsets of
\(\mathbb R^{d_h}\),
\begin{equation}
  -\tau_h B_\xi^{(K)}(\iota_c(y)) = \tau_h\langle \xi,y\rangle + O(c), \qquad c\to0^+.
\end{equation}
Consequently, fixed-temperature HSA converges to Euclidean
\emph{directional-projection attention}.  If the fixed temperature is replaced
by a query-dependent scale
\(\beta_i=\|q_i\|/\sqrt{d_h}\) and \(\xi_i=q_i/\|q_i\|\), then the limiting
score is the usual scaled dot product
\begin{equation}
  \beta_i\langle \xi_i,y_j\rangle
  =
  \frac{\langle q_i,y_j\rangle}{\sqrt{d_h}} .
\end{equation}
\end{proposition}

\begin{proposition}[Radial--angular score decomposition]
\label{prop:decompose}
Let \(k=(k_0,k_s)\in\mathbb L_K^{d_h}\), let
\(\xi\in\mathbb S^{d_h-1}\), and write
\(p=k_s/k_0\in B^{d_h}\) for the Klein coordinate of \(k\). Then
\begin{equation}
  -\tau_h B_\xi^{(K)}(k)
  =
  \frac{\tau_h}{c}
  (
    \underbrace{-\log(ck_0)}_{\text{radial term}}
    +
    \underbrace{\log\frac{1}{1-\langle p,\xi\rangle}}_{\text{angular alignment}}
  ).
  \label{eq:score-decomp-main}
\end{equation}
Since \(ck_0\ge 1\), the radial term is non-positive and decreases as the key
moves away from the basepoint. The angular term rewards alignment with the query-selected boundary direction
\(\xi\), favouring keys on the selected asymptotic branch. Thus the directed HSA score couples boundary alignment with the key's position in the
hyperboloid.

At \(K=-1\), multiplying this decomposition by the query-dependent factor in
\modelld{} gives two additional interaction terms. With
\(p_j=k_{j,s}/k_{j,0}\),
\begin{align}
  s_{ij}^{\mathrm{ld}}
  ={}&
  \underbrace{-\tau_h\log k_{j,0}}_{\text{key radial term}}
  +
  \underbrace{\tau_h\log\frac{1}{1-\langle p_j,\xi\rangle}}_{\text{angular alignment}}
  -
  \underbrace{\tau_h\gamma_h\log q_{i,0}\log k_{j,0}}_{\text{query--key radial coupling}}
  +
  \underbrace{\tau_h\gamma_h\log q_{i,0}
  \log\frac{1}{1-\langle p_j,\xi\rangle}}_{\text{query-depth-gated alignment}}.
  \label{eq:ld-decompose}
\end{align}
The last two terms vanish when \(\gamma_h=0\). For deeper queries
\((\log q_{i,0}>0)\), the third term strengthens the key-depth penalty, while
the fourth term strengthens the boundary-alignment reward. Thus log-depth
modulation sharpens the contrast between poorly aligned deep keys and keys
well aligned with the selected boundary direction.
\end{proposition}

\section{Experiments}
\label{sec:experiments}

We evaluate \modelname{} on node classification and link prediction
across several graph benchmarks, then study attention-score design,
the GNN branch, and the computational cost of linear boundary attention.

\subsection{Datasets and Protocols}
\label{sec:datasets}

We use Disease and Airport as hierarchical graph benchmarks and Cora,
Citeseer, and PubMed as citation benchmarks. Node classification reports
macro-F1 on Disease and accuracy elsewhere over ten seeds. Large-scale
classification uses ogbn-arxiv~\citep{hu2020ogb} with its official chronological
split. Link prediction reports ROC-AUC on Disease,
Airport, PubMed, and Cora.

\subsection{Baselines}
\label{sec:baselines}

For node classification, we compare against GCN~\citep{kipf2017gcn},
GAT~\citep{velickovic2018gat} (Euclidean GNNs), HGCN~\citep{chami2019hgcn},
Hypformer~\citep{yang2024hypformer}, and a Euclidean Transformer. Large-scale comparisons
also include SGFormer~\citep{wu2023sgformer} and the official OGB GCN.
Implementation and reporting details are in
Appendix~\ref{app:experimental-reporting}.

\subsection{Node Classification Results}
\label{sec:nc-results}

\begin{table}[htbp]
\centering
\caption{Node classification: mean $\pm$ population SD (\%) over ten
seeds. Disease uses macro-F1; other datasets use accuracy.
Bold denotes the highest mean. Implementation details and splits are given in
Appendix~\ref{app:impl}.}
\label{tab:nc}
\footnotesize
\setlength{\tabcolsep}{3pt}
\begin{tabular}{lccccc}
\toprule
Model & Disease & Airport & Cora & Citeseer & PubMed \\
\midrule
GCN
& $79.57\pm3.65$ & $83.87\pm2.32$ & $78.67\pm1.74$
& $68.97\pm1.25$ & $71.55\pm1.37$ \\
GAT
& $78.31\pm4.81$ & $85.61\pm2.07$ & $79.39\pm1.23$
& $68.86\pm1.26$ & $76.02\pm1.48$ \\
HGCN
& $81.75\pm5.77$ & $83.81\pm2.16$ & $79.58\pm0.85$
& $67.07\pm1.02$ & $76.35\pm0.46$ \\
Hypformer
& $78.98\pm3.95$ & $\mathbf{92.49\pm1.27}$ & $72.34\pm1.67$
& $64.98\pm1.12$ & $71.62\pm1.40$ \\
Euclidean Transformer
& $90.49\pm2.43$ & $92.15\pm1.94$ & $76.58\pm1.16$
& $66.48\pm1.40$ & $75.31\pm0.75$ \\
\midrule
\modelld{}
& $\mathbf{92.66\pm3.16}$ & $91.53\pm1.40$ & $\mathbf{80.94\pm0.73}$
& $\mathbf{69.19\pm0.96}$ & $\mathbf{78.48\pm0.48}$ \\
\bottomrule
\end{tabular}
\end{table}

\modelld{} achieves the highest mean on Disease, Cora, Citeseer, and
PubMed (Table~\ref{tab:nc}), exceeding the strongest displayed baseline
by 2.17, 1.36, 0.22, and 2.13 pp, respectively. On Airport, it reaches
91.53\% accuracy, compared with 92.49\% for Hypformer.

\FloatBarrier
\subsection{Link Prediction Results}
\label{sec:lp-results}

\begin{table}[htbp]
\centering
\caption{Link prediction (test ROC-AUC, \%). Use a squared-geodesic Fermi--Dirac decoder ($p=2$).}
\label{tab:lp}
\small
\setlength{\tabcolsep}{4pt}
\begin{tabular}{lcccc}
\toprule
Method & Disease-LP & Airport-LP & PubMed-LP & Cora-LP \\
\midrule
\multicolumn{5}{l}{\textit{Euclidean GNNs}} \\
GCN~\citep{kipf2017gcn}
& $81.24\pm0.68$ & $89.3\pm0.4$ & $91.1\pm0.5$ & $90.4\pm0.2$ \\
GAT~\citep{velickovic2018gat}
& $76.37\pm1.19$ & $90.5\pm0.3$ & $91.2\pm0.1$ & $93.7\pm0.1$ \\
\midrule
\multicolumn{5}{l}{\textit{Hyperbolic GNNs}} \\
HGCN~\citep{chami2019hgcn}
& $90.8\pm0.3$ & $96.4\pm0.1$ & $96.3\pm0.0$ & $92.9\pm0.1$ \\
HBNN~\citep{chen2026hbnn}
& $84.90\pm0.89$ & $94.88\pm0.39$ & $94.39\pm0.27$ & $91.42\pm0.68$ \\
\midrule
\multicolumn{5}{l}{\textit{BusemannFormer variants}} \\
\modelld{} (directed, $p=2$)
& $89.99\pm1.91$ & $93.01\pm0.42$ & $91.92\pm0.30$ & $89.00\pm0.85$ \\
\modelld{} (bidirectional)& $90.16\pm1.52$ & $93.06\pm0.57$ & $90.71\pm0.43$ & $90.19\pm0.49$ \\
\bottomrule
\end{tabular}
\end{table}

\paragraph{Score and decoder choice.}
The LD variants in Table~\ref{tab:lp} use log-depth Busemann
attention and a squared-geodesic Fermi--Dirac decoder,
$\sigma((r-d_{\mathbb H}(u,v)^2)/t)$, where $t$ is the effective
positive decoder temperature.

Overall, \modelld{} is competitive with HGCN on Disease and compares
favorably with GCN/GAT on PubMed, while stronger baselines remain ahead
on Airport and Cora.
\subsection{Ablation Studies}
\label{sec:ablation}

\paragraph{Attention Score Design.}
Table~\ref{tab:ablation-score} reports five score controls and the selected
LD results from Table~\ref{tab:nc}. The controls hold all settings other
than the attention score fixed and disable the GNN. Write
$b(q,k)=B_{\xi(q)}(k)$ and
$g(q)=1+\gamma\log\max\{q_0,1+\epsilon\}$, omitting head indices.

\begin{table}[htbp]
\centering
\caption{Attention-score ablation: mean $\pm$ population SD (\%) over ten
seeds; macro-F1 on Disease and accuracy elsewhere. Fixed-architecture
controls, with their highest means in bold.}
\label{tab:ablation-score}
\footnotesize
\setlength{\tabcolsep}{1.5pt}
\begin{tabular}{@{}llccccc@{}}
\toprule
Score function & Formula & Disease & Airport & Cora & Citeseer & PubMed \\
\midrule
Log-depth HSA
& $-\tau g(q)b(q,k)$
& $\mathbf{92.66}\pm3.16$ & $91.53\pm1.40$ & $75.37\pm1.43$
& $66.67\pm1.63$ & $76.31\pm0.88$ \\
Hyperbolic distance
& $-\tau d_{\mathbb H}(q,k)^2$
& $91.20\pm2.91$ & $92.01\pm1.28$ & $77.02\pm0.85$
& $66.65\pm1.49$ & $76.39\pm0.82$ \\
Spatial dot product
& $\tau q_s^\top k_s$
& $91.87\pm2.35$ & $92.28\pm0.53$ & $76.48\pm0.92$
& $\mathbf{67.47}\pm1.01$ & $76.71\pm1.08$ \\
Bidirectional HSA
& $-\frac{\tau}{2}[b(q,k)+b(k,q)]$
& $90.03\pm3.43$ & $91.36\pm1.01$ & $77.34\pm0.87$
& $66.94\pm1.23$ & $76.37\pm0.91$ \\
Directed HSA
& $-\tau b(q,k)$
& $90.17\pm2.48$ & $\mathbf{92.30}\pm1.30$ & $\mathbf{77.37}\pm0.85$
& $66.95\pm1.19$ & $\mathbf{76.79}\pm0.89$ \\
\bottomrule
\end{tabular}
\end{table}

Among the controls, LD has the highest Disease mean, spatial-dot leads on
Citeseer, and directed HSA leads on Airport, Cora, and PubMed. On Airport,
directed HSA and spatial-dot scoring have nearly identical means
(92.30\% and 92.28\%, respectively). These results show that HSA variants
outperform or are competitive with other hyperbolic attention scores.

\paragraph{Effect of the GNN Branch.}
\label{sec:ablation-arch}
Table~\ref{tab:gnn-branch} separates the Lorentz GNN, log-depth HSA,
and their combination. Adding HSA to the GNN improves the mean on all
five datasets by 1.21--6.09 pp. HSA alone performs best on Disease and
Airport, whereas the combined model performs best on Cora, Citeseer,
and PubMed. These results indicate complementary, dataset-dependent
contributions; the performance cannot be attributed to the GNN alone.

\begin{table}[htbp]
\centering
\caption{Component ablation: ten-seed test means
(\%), using macro-F1 on Disease and accuracy elsewhere. Protocol details are in Appendix~\ref{app:arch-ablation}.}
\label{tab:gnn-branch}
\small
\setlength{\tabcolsep}{5pt}
\begin{tabular}{lccccc}
\toprule
Architecture & Disease & Airport & Cora & Citeseer & PubMed \\
\midrule
Lorentz GNN only & 84.76 & 58.47 & 75.96 & 65.57 & 75.35 \\
Log-depth HSA only & 92.66 & 61.49 & 75.37 & 66.67 & 76.31 \\
Log-depth HSA + GNN & 90.85 & 60.90 & 77.17 & 66.86 & 77.37 \\
\bottomrule
\end{tabular}
\end{table}

\paragraph{Linearization Analysis.}
\label{sec:large-graph-results}

We evaluate scalable Busemann-based attention on ogbn-arxiv
(Table~\ref{tab:linear-performance}). Busemann modulation of Hypformer's
native linear attention reaches $72.912\pm0.114\%$ test accuracy,
compared with $72.946\pm0.116\%$ for unmodified Hypformer.
The boundary-dictionary extension is evaluated separately in the lower
block, alongside a focused-positive attention control. These smaller
models reach 69.878--69.979\% and use fewer parameters, but trail the
external baselines. Increasing the dictionary from 16 to 64 directions
does not establish an accuracy improvement.

\begin{table}[htbp]
\centering
\caption{ogbn-arxiv: test accuracy (mean $\pm$ sample SD, five seeds),
parameters, and H200 inference cost. Busemann modulation modifies native
Hypformer attention rather than replacing it with a boundary dictionary.
The lower block shares the Busemann
backbone; external models use their own recipes.}
\label{tab:linear-performance}
\footnotesize
\setlength{\tabcolsep}{4pt}
\begin{tabular}{lrrrr}
\toprule
Model / global kernel & Test accuracy (\%) & Parameters & ms & MiB \\
\midrule
Hypformer-large, native
& $\mathbf{72.946\pm0.116}$ & 848,723 & 39.97 & 4,467.0 \\
Hypformer + Busemann modulation
& $72.912\pm0.114$ & 979,797 & 55.09 & 5,128.5 \\
SGFormer, official rerun
& $72.747\pm0.281$ & 474,152 & 20.05 & 1,314.2 \\
GCN, official OGB rerun
& $71.804\pm0.095$ & 110,120 & 9.13 & 961.2 \\
\midrule
\modelname{}, dictionary 16
& $69.878\pm0.315$ & 113,202 & 59.55 & 1,273.2 \\
Same backbone, focused 16
& $69.885\pm0.269$ & 111,018 & 61.09 & 1,273.2 \\
Same backbone, dictionary 64
& $69.979\pm0.266$ & 119,730 & 64.31 & 1,273.2 \\
\bottomrule
\end{tabular}
\end{table}

\FloatBarrier
\subsection{Qualitative Analysis}
\label{sec:qualitative}

\begin{figure}[htbp]
\centering
\includegraphics[width=\linewidth]{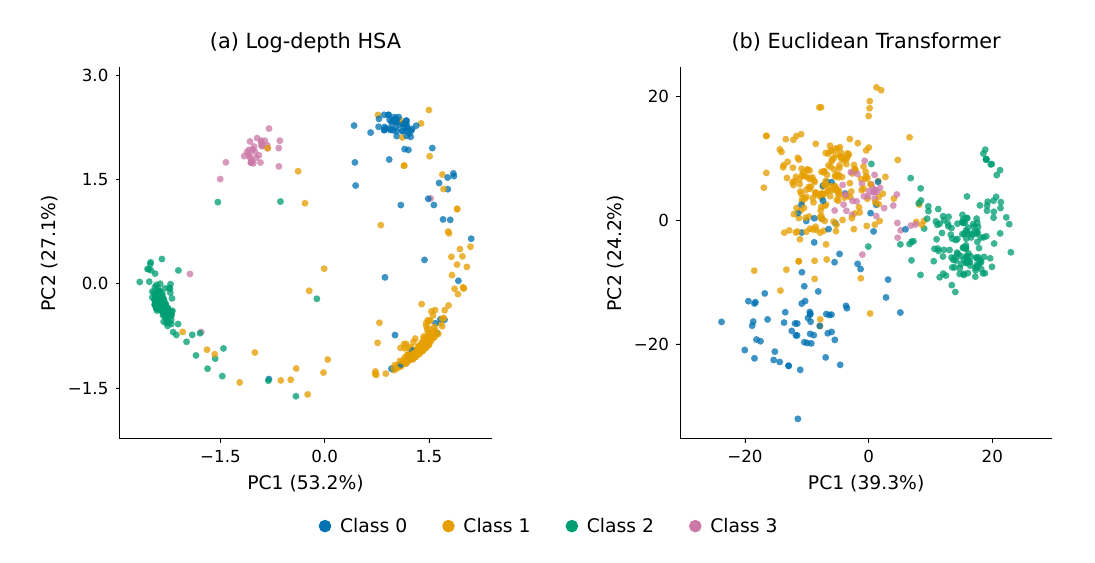}
\caption{Airport test-node representations for \modelld{} and the Euclidean
Transformer ($D=64,H=4,L=2$, no GNN; seed 51). Colors denote the four classes; axes
report explained training variance. Both panels show the same 478
test nodes.}
\label{fig:visualization}
\end{figure}

Figure~\ref{fig:visualization} visualizes the learned representations of
478 Airport test nodes using PCA. Under \modelld{}, nodes form
class-dominated regions along a curved configuration, with particularly
distinct groups for classes 2 and 3 and some overlap between classes 0
and 1. The Euclidean Transformer also exhibits class structure, although
classes 1 and 3 overlap more in this projection. These patterns illustrate 
class-relevant information in the learned representations of BusemannFormer.
Projection details are given in Appendix~\ref{app:qualitative}.

\section{Related Work}
\label{sec:related}

\paragraph{Hyperbolic representation learning.}
Hyperbolic geometry is well suited for hierarchical and complex-network
data~\citep{nickel2017poincare,nickel2018lorentz,sala2018representation,
krioukov2010hyperbolic}. Hyperbolic neural layers have been developed in the
Poincar\'e and Lorentz models~\citep{ganea2018hnn,shimizu2021hnn,
chen2022fully,zhang2021lorentzian,bachmann2020kappagcn}, with extensions to
graph convolutions~\citep{chami2019hgcn,liu2019hgnn} and image
tasks~\citep{khrulkov2020hyperbolic}; see~\citet{peng2022survey} for a
survey. HBNN~\citep{chen2026hbnn} unified hyperbolic layers through Busemann
functions, introducing the BFC and BMLR primitives; we use BMLR for node
classification and HypLinear for projections.

\paragraph{Graph Transformers.}
Graphormer~\citep{ying2021graphformer} injects graph structure through
centrality and shortest-path encodings, while GraphGPS~\citep{rampavsek2022recipe}
combines message passing with Transformer layers. NodeFormer~\citep{wu2022nodeformer}
and SGFormer~\citep{wu2023sgformer} improve scalability to $O(N)$ complexity.
These methods are Euclidean and do not exploit hyperbolic geometry.

\paragraph{Hyperbolic Transformers.}
Building on early hyperbolic attention~\citep{gulcehre2019hyperbolic},
Hypformer~\citep{yang2024hypformer} proposed a fully Lorentz-model
hyperbolic Transformer with linear self-attention. Our difference is the
\emph{attention score}: Hypformer uses a hyperbolic dot-product-like score,
whereas we use the query-dependent Busemann depth
$-\tau B_{v_i}(k_j)$, allowing each token to choose an ideal-boundary
direction and capture asymmetric depth relationships
(Table~\ref{tab:comparison}). HyperET~\citep{hyperET2025},
HypLoRA~\citep{hyplora2025}, and HELM~\citep{helm2025} apply hyperbolic
geometry to LLMs rather than graph-structured data.

\paragraph{Busemann and horospherical methods.}
Busemann functions have been used for hyperbolic PCA~\citep{chami2021horopca},
ideal-boundary prototypes~\citep{ghadimi2021busemann}, large-margin
classification~\citep{fan2023horosphere}, and fixed class-level energy
attention~\citep{kapic2026embolic}. These methods use fixed or class-level
boundary directions. In contrast, \modelname{} lets each query token select
its own boundary direction at inference time, yielding a general
self-attention mechanism rather than classification-specific scoring.

A detailed conceptual comparison with Hypformer, HBNN, Busemann prototypes,
and EmBolic is given in Appendix~\ref{app:comparison}
(Table~\ref{tab:comparison}).
\section{Conclusion}
\label{sec:conclusion}

We introduced \modelname{}, a hyperbolic Transformer built on
Horospherical Self-Attention, which uses the ideal boundary to make
directional depth explicit in token-to-token retrieval. Our theoretical
analysis characterizes its geometric properties and connection to
Euclidean attention. Experiments demonstrate strong node-classification
performance and competitive, dataset-dependent link-prediction results,
while controlled ablations clarify the contributions of attention
scoring and graph aggregation. A finite-boundary-dictionary extension
further enables linear-complexity global attention through a modified
kernel. Together, these findings establish boundary-directed attention
as a promising approach to learning with hierarchical data.

\paragraph{Limitations.}
Unrestricted HSA remains quadratic. The boundary-dictionary extension
achieves linear complexity for a fixed dictionary size but changes
the original attention kernel. Moreover, our large-scale predictive
evaluation is limited to ogbn-arxiv. Broader evaluation on larger and
more structurally diverse graphs is needed to characterize the
extension's accuracy--efficiency trade-off and establish its practical
advantages over existing scalable attention mechanisms.

\clearpage
\appendix

\section{Hyperbolic Geometry}
\label{app:hyperbolic}

We collect the geometric definitions and identities used throughout
the paper, writing the negative curvature as \(K=-c^2<0\).

\subsection{The Lorentz hyperboloid model}
\label{app:lorentz}

\paragraph{Lorentz inner product.}
We write a point $x\in\mathbb{R}^{d+1}$ in coordinates $(x_0,x_s)$,
where $x_0\in\mathbb{R}$ and $x_s\in\mathbb{R}^d$.
Let $\langle\cdot,\cdot\rangle$ and $\|\cdot\|$ denote the standard
Euclidean inner product and norm on $\mathbb{R}^d$.
For $x=(x_0,x_s)$ and $y=(y_0,y_s)$, the Lorentz inner product is
\begin{equation}
  \langle x,y\rangle_{\mathbb{L}} = -x_0 y_0 + \langle x_s,y_s\rangle.
\end{equation}
We also write $\|u \|_{\mathbb{L}}=\sqrt{\langle u,u\rangle_{\mathbb{L}}}$.

\paragraph{Hyperboloid and distance.}
The Lorentz model of \(d\)-dimensional hyperbolic space with curvature
\(K=-c^2\) is
\begin{equation}
  \mathbb L_K^d
  =
  \{x\in\mathbb R^{d+1}:\langle x,x\rangle_{\mathbb L}=-c^{-2},\;x_0>0\}.
\end{equation}
The basepoint is \(o_K=(c^{-1},0)\).  For \(x,y\in\mathbb L_K^d\),
\begin{equation}
  d_{\mathbb{H}}(x,y)
  =
  \frac1c\operatorname{arcosh}\!\left(-c^2\langle x,y\rangle_{\mathbb L}\right).
\end{equation}
Equivalently,
\(-c^2\langle x,y\rangle_{\mathbb L}=\cosh(c\,d_{\mathbb{H}}(x,y))\).
\paragraph{Tangent spaces, exponential map, and logarithmic map.}
The tangent space at \(x\in\mathbb L_K^d\) is
\begin{equation}
  T_x\mathbb L_K^d
  =
  \{u\in\mathbb R^{d+1}:\langle x,u\rangle_{\mathbb L}=0\}.
\end{equation}
For \(u\in T_x\mathbb L_K^d\), let \(r=\|u\|_x\).  The exponential map is
\begin{equation}
  \operatorname{Exp}_x^K(u)
  =
  \cosh(cr)x
  +
  \frac{\sinh(cr)}{cr}u,
  \label{eq:expmap-correct}
\end{equation}
with the usual continuous interpretation when \(r=0\).  If
\(y\in\mathbb L_K^d\) and \(\alpha=-c^2\langle x,y\rangle_{\mathbb L}\), then
\begin{equation}
  \operatorname{Log}_x^K(y)
  =
  \frac{\operatorname{arcosh}(\alpha)}{\sqrt{\alpha^2-1}}
  \left(y+c^2\langle x,y\rangle_{\mathbb L}x\right),
  \label{eq:logmap-correct}
\end{equation}
again with the continuous value \(0\) when \(x=y\).  The vector in parentheses
is tangent to \(\mathbb L_K^d\) at \(x\).

For a unit vector \(\xi\in\mathbb S^{d-1}\), the unit-speed geodesic from
\(o_K\) in direction \(\xi\) is
\begin{equation}
  \gamma_\xi(t)
  =
  \operatorname{Exp}_{o_K}^K(0,t\xi)
  =
  \left(\frac1c\cosh(ct),\frac1c\sinh(ct)\xi\right).
  \label{eq:unit-geodesic-correct}
\end{equation}
At \(K=-1\), this reduces to \((\cosh t,\sinh t\,\xi)\).

\subsection{Ideal boundary and Busemann functions}
\label{app:busemann}

The ideal boundary \(\partial_\infty\mathbb L_K^d\) is the set of asymptotic
classes of geodesic rays.  With the basepoint \(o_K\) fixed, it is identified
with \(\mathbb S^{d-1}\) by \(\xi\mapsto[\gamma_\xi]\).

For \(\xi\in\mathbb S^{d-1}\), define the Busemann function by
\begin{equation}
  B_\xi^{(K)}(x)
  =
  \lim_{t\to\infty}\bigl(d_{\mathbb{H}}(x,\gamma_\xi(t))-t\bigr).
\end{equation}
Using~\eqref{eq:unit-geodesic-correct} and
\(-c^2\langle x,\gamma_\xi(t)\rangle_{\mathbb L}
=c(x_0\cosh(ct)-\langle x_s,\xi\rangle\sinh(ct))\), this limit gives
\begin{equation}
  B_\xi^{(K)}(x)
  =
  \frac1c\log\!\left(c(x_0-\langle x_s,\xi\rangle)\right).
  \label{eq:buse-lorentz-correct}
\end{equation}
Indeed, as \(t\to\infty\),
\begin{equation}
  d_{\mathbb{H}}(x,\gamma_\xi(t))
  =
  t+\frac1c\log\!\left(c(x_0-\langle x_s,\xi\rangle)\right)+o(1).
\end{equation}
The Busemann function has the following properties.
\begin{enumerate}[leftmargin=*]
  \item \(B_\xi^{(K)}(o_K)=0\);
  \item \(B_\xi^{(K)}(\gamma_\xi(t))=-t\);
  \item \(B_\xi^{(K)}\) is 1-Lipschitz and
        \(\|\nabla^{\mathbb L}B_\xi^{(K)}\|=1\);
  \item \(x_0-\langle x_s,\xi\rangle>0\) for every \(x\in\mathbb L_K^d\).
\end{enumerate}

\paragraph{Flat limit in normal coordinates.}
For \(y\in\mathbb R^d\) and \(r=\|y\|\), the normal-coordinate embedding is
\begin{equation}
    \iota_c(y)=\operatorname{Exp}_{o_K}^K(0,y) =
  \left(\frac1c\cosh(cr),\frac1c\frac{\sinh(cr)}{r}y\right)~.
\end{equation} By direct computation,
\begin{align}
  c\bigl(\iota_c(y)_0-\langle \iota_c(y)_s,\xi\rangle\bigr)
  &=
  \cosh(cr)-\frac{\sinh(cr)}{r}\langle y,\xi\rangle \\
  &=
  1-c\langle y,\xi\rangle+O(c^2),
\end{align}
so
\begin{equation}
  B_\xi^{(K)}(\iota_c(y))
  =
  -\langle y,\xi\rangle+O(c),
  \qquad c\to0^+.
\end{equation}
This is the precise sense in which \(-B_\xi^{(K)}\) limits to a Euclidean
linear functional.

\subsection{Horospheres}
\label{app:horospheres}

For \(\xi\in\mathbb S^{d-1}\) and \(\alpha\in\mathbb R\), the horosphere
centred at \(\xi\) with Busemann value \(\alpha\) is
\begin{equation}
  H_{\xi,\alpha}^{(K)}
  =
  \{x\in\mathbb L_K^d:B_\xi^{(K)}(x)=\alpha\}.
\end{equation}
The family \(\{H_{\xi,\alpha}^{(K)}\}_{\alpha\in\mathbb R}\) foliates
\(\mathbb L_K^d\).  Horospheres are intrinsically flat and are the hyperbolic
analogues of Euclidean affine hyperplanes orthogonal to the direction \(\xi\).
For fixed centre \(\xi\), the signed distance between
\(H_{\xi,\alpha_1}^{(K)}\) and \(H_{\xi,\alpha_2}^{(K)}\) along the normal
geodesics is \(\alpha_2-\alpha_1\) and the unsigned distance is
\(|\alpha_2-\alpha_1|\). Appendix~\ref{app:horosphere-levels} identifies HSA iso-score sets with these horospheres.

\subsection{Fr\'echet mean and Lorentz centroid}
\label{app:frechet}

For a probability measure \(\mu\) on \(\mathbb L_K^d\), its Fr\'echet mean is
\begin{equation}
  \bar x
  =
  \operatorname*{argmin}_{x\in\mathbb L_K^d}
  \int d_{\mathbb{H}}(x,y)^2\,d\mu(y).
\end{equation}
Since \(\mathbb L_K^d\) is complete, simply connected, and nonpositively
curved, the Fr\'echet mean of a finite measure with finite second moment is
unique.

For nonnegative weights \(a_j\) with \(\sum_j a_j=1\), the ambient weighted
sum \(m=\sum_j a_jx_j\) of future-pointing hyperboloid points is future
timelike whenever it is nonzero.  The Lorentz centroid projection is
\begin{equation}
  o
  =
  \frac{m}{c\sqrt{-\langle m,m\rangle_{\mathbb L}}}
  \in\mathbb L_K^d.
  \label{eq:lorentz-centroid-correct}
\end{equation}
This gives the Lorentz centroid used as an efficient approximation to the Fr\'echet mean.

\section{Proofs and Auxiliary Facts}
\label{app:proofs}

This appendix proves the two propositions stated in Section~\ref{sec:theory}
and records the auxiliary geometric facts used elsewhere in the paper.  We use
the convention \(K=-c^2\), \(c>0\), and
\[
  B_\xi^{(K)}(x)
  =
  \frac1c\log\!\left(c(x_0-\langle x_s,\xi\rangle)\right).
\]

\subsection{Proof of Proposition~\ref{prop:euclidean-limit}}
\label{app:proof-flat-limit}

Let \(r=\|y\|\).  By the curvature-\(K\) exponential map at
\(o_K=(c^{-1},0)\),
\begin{equation}
  \iota_c(y)
  =
  \operatorname{Exp}_{o_K}^{K}(0,y)
  =
  \left(\frac1c\cosh(cr),\frac1c\frac{\sinh(cr)}{r}y\right),
\end{equation}
with the continuous interpretation at \(r=0\).  Hence
\begin{align}
  c\bigl(\iota_c(y)_0-\langle \iota_c(y)_s,\xi\rangle\bigr)
  &=
  \cosh(cr)-\frac{\sinh(cr)}{r}\langle y,\xi\rangle \\
  &=
  1-c\langle y,\xi\rangle+O(c^2),
\end{align}
where the error is uniform for \(y\) in compact subsets of \(\mathbb R^{d_h}\).
Therefore
\begin{align}
  B_\xi^{(K)}(\iota_c(y))
  &=
  \frac1c\log\left(1-c\langle y,\xi\rangle+O(c^2)\right) \\
  &=
  -\langle y,\xi\rangle+O(c).
\end{align}
Multiplying by \(-\tau_h\) proves the fixed-temperature limit.  If
\(\xi_i=q_i/\|q_i\|\) and \(\beta_i=\|q_i\|/\sqrt{d_h}\), then
\[
  \beta_i\langle \xi_i,y_j\rangle
  =
  \frac{\langle q_i,y_j\rangle}{\sqrt{d_h}},
\]
which is the scaled dot-product score.

\subsection{Proof of Proposition~\ref{prop:decompose}}
\label{app:proof-decompose}

Since \(p=k_s/k_0\),
\begin{equation}
  c(k_0-\langle k_s,\xi\rangle)
  =
  ck_0\bigl(1-\langle p,\xi\rangle\bigr).
\end{equation}
The hyperboloid constraint implies \(\|p\|<1\), hence
\(1-\langle p,\xi\rangle>0\).  Taking logarithms gives
\begin{align}
  -\tau_hB_\xi^{(K)}(k)
  &=-\frac{\tau_h}{c}
  \log\!\left(ck_0(1-\langle p,\xi\rangle)\right) \\
  &=
  \frac{\tau_h}{c}
  \left[-\log(ck_0)+\log\frac{1}{1-\langle p,\xi\rangle}\right].
\end{align}
Moreover \(ck_0\ge1\), so \(-\log(ck_0)\le0\).  The sign of the angular term is
the sign of \(\langle p,\xi\rangle\): it is positive for positive alignment,
zero at orthogonality, and negative under anti-alignment.

At \(K=-1\), equation~\eqref{eq:ld-decompose} follows by multiplying the
\(K=-1\) decomposition by \((1+\gamma_h\log q_{i,0})\) and collecting the four
terms.  Since \(q_{i,0}\ge1\), the query-depth factor \(\log q_{i,0}\) is
non-negative.

\subsection{Horosphere level sets}
\label{app:horosphere-levels}

For fixed query direction \(v_i^{(h)}\) and temperature \(\tau_h>0\),
\begin{equation}
  s_{ij}^{(h)}=\alpha
  \quad\Longleftrightarrow\quad
  B_{v_i^{(h)}}^{(K)}(k_j^{(h)})=-\alpha/\tau_h.
\end{equation}
Thus each iso-score set is the horosphere
\begin{equation}
  H_{v_i^{(h)},-\alpha/\tau_h}^{(K)}
  =
  \{k\in\mathbb L_K^{d_h}:B_{v_i^{(h)}}^{(K)}(k)=-\alpha/\tau_h\}
\end{equation}
centred at the query-selected ideal-boundary point \(v_i^{(h)}\).  This is the
exact hyperbolic analogue of Euclidean dot-product attention having affine
hyperplane level sets.

\subsection{Gradient stability}
\label{app:stability}

Let
\begin{equation}
  u_{ij}^{(h)}
  =
  c\bigl(k_{j,0}^{(h)}-\langle k_{j,s}^{(h)},v_i^{(h)}\rangle\bigr).
\end{equation}
If \(u_{ij}^{(h)}\ge\delta>0\), then the Euclidean ambient-coordinate gradient
of \(s_{ij}^{(h)}\) with respect to the key coordinates satisfies
\begin{equation}
  \|\nabla_{k_j}^{\mathrm{amb}}s_{ij}^{(h)}\|_2
  \le
  \frac{\tau_h\sqrt2}{\delta}.
\end{equation}
Indeed, suppressing indices and writing \(u=c(k_0-\langle k_s,v\rangle)\),
\[
  s=-\frac{\tau_h}{c}\log u,
  \qquad
  \frac{\partial s}{\partial k_0}=-\frac{\tau_h}{u},
  \qquad
  \frac{\partial s}{\partial k_s}=\frac{\tau_h v}{u}.
\]
Since \(\|v\|=1\), the ambient norm is
\[
  \|\nabla_k^{\mathrm{amb}}s\|_2^2
  =
  \frac{\tau_h^2}{u^2}(1+\|v\|^2)
  =
  \frac{2\tau_h^2}{u^2}
  \le
  \frac{2\tau_h^2}{\delta^2}.
\]
Intrinsically, Busemann functions satisfy \(\|\nabla^{\mathbb L}B_\xi^{(K)}\|=1\),
so \(\|\nabla^{\mathbb L}(-\tau_hB_\xi^{(K)})\|=\tau_h\).  The ambient bound is
therefore an implementation-level bound for coordinate computations and
clamping, not a claim that the intrinsic Busemann gradient becomes large.

\subsection{Geodesic depth ordering and branch separation}
\label{app:geodesic-depth}

Let \(\gamma_\eta(t)\) be the unit-speed ray from \(o_K\) toward
\(\eta\in\mathbb S^{d_h-1}\):
\begin{equation}
  \gamma_\eta(t)
  =
  \left(\frac1c\cosh(ct),\frac1c\sinh(ct)\eta\right).
\end{equation}
For \(a=\langle \xi,\eta\rangle\), direct substitution gives
\begin{equation}
  B_\xi^{(K)}(\gamma_\eta(t))
  =
  \frac1c\log\bigl(\cosh(ct)-a\sinh(ct)\bigr).
  \label{eq:branch-busemann-app}
\end{equation}
If \(\xi=\eta\), then \(a=1\) and
\[
  B_\xi^{(K)}(\gamma_\xi(t))
  =
  \frac1c\log(e^{-ct})=-t,
  \qquad
  -\tau_hB_\xi^{(K)}(\gamma_\xi(t))=\tau_h t.
\]
Thus the score grows linearly with geodesic depth, with slope \(\tau_h\), for
all curvatures \(K=-c^2\) when \(t\) is distance.

If \(\xi\ne\eta\), then \(a<1\) and
\begin{equation}
  \cosh(ct)-a\sinh(ct)
  =
  \frac{1-a}{2}e^{ct}+\frac{1+a}{2}e^{-ct}.
\end{equation}
Hence
\begin{equation}
  B_\xi^{(K)}(\gamma_\eta(t))
  =
  t+\frac1c\log\left(\frac{1-a}{2}\right)+o(1),
  \qquad t\to\infty,
\end{equation}
and therefore
\begin{equation}
  -\tau_hB_\xi^{(K)}(\gamma_\eta(t))
  =
  -\tau_h t
  -
  \frac{\tau_h}{c}\log\left(\frac{1-\langle \xi,\eta\rangle}{2}\right)
  +o(1).
\end{equation}
The score gap between the selected branch \(\xi\) and a distinct branch
\(\eta\) grows as \(2\tau_h t+O(1)\).  This is the branch-separation property
used to interpret HSA as an asymptotic branch selector.

\subsection{Bidirectional score and conditional symmetry}
\label{app:bidirectional-symmetry}

The implemented \modelsym{} score is better viewed as a bidirectional
Busemann score:
\begin{equation}
  s_{ij}^{(h),\mathrm{bi}}
  =
  -\frac{\tau_h}{2}
  \left(
    B_{v_i^{Q,(h)}}^{(K)}(k_j^{(h)})
    +
    B_{v_j^{K,(h)}}^{(K)}(q_i^{(h)})
  \right).
\end{equation}
If the query and key maps are tied, so that \(q_i^{(h)}=k_i^{(h)}=y_i^{(h)}\)
and the boundary direction attached to token \(i\) is the same in both terms,
then
\begin{equation}
  s_{ij}^{(h),\mathrm{bi}}
  =
  -\frac{\tau_h}{2}
  \left(
    B_{v_i^{(h)}}^{(K)}(y_j^{(h)})
    +
    B_{v_j^{(h)}}^{(K)}(y_i^{(h)})
  \right)
  =
  s_{ji}^{(h),\mathrm{bi}}.
\end{equation}
Without tied query/key maps this equality need not hold.  Thus \modelsym{} is
symmetric as a mathematical score only in the tied-map case; otherwise it is a
bidirectional depth score that uses both query-to-key and key-to-query
Busemann evaluations.

\section{Implementation Details}
\label{app:impl}
\label{app:exp}

\subsection{Hyperparameter Grid}

\begin{table}[h]
\centering
\caption{Reference hyperparameter ranges and defaults.}
\label{tab:hparam}
\small
\begin{tabular}{lll}
\toprule
\textbf{Hyperparameter} & \textbf{Range} & \textbf{Default} \\
\midrule
Hidden dim $d$ & \{32, 64, 128\} & 64 \\
Heads $H$ & \{2, 4, 8\} & 4 \\
Layers $L$ & \{1, 2, 3\} & 2 \\
Learning rate & \{1e-4, 5e-4, 1e-3\} & 5e-4 \\
Weight decay & \{0, 1e-4, 1e-3\} & 1e-4 \\
Dropout & \{0.0, 0.1, 0.2\} & 0.1 \\
$K$ (hop mask) & \{1, 2, 3\} & 2 \\
Temperature $\tau_h$ init & $\sqrt{d_h}$ & learnable \\
\bottomrule
\end{tabular}
\end{table}

\subsection{Numerical Stability}
\label{sec:numerical}

To prevent $k_{j,0}-\langle k_{j,s},v_i\rangle$ from approaching zero:
(1) we clamp the argument to $[\varepsilon, \infty)$ with $\varepsilon=10^{-6}$;
(2) we apply gradient clipping with max-norm 1;
(3) optionally, a radial penalty $\lambda\sum_j \|k_{j,s}\|^2$
encourages points to remain far from the ideal boundary.

\subsection{Dataset Partitions}
\label{app:dataset-partitions}

Table~\ref{tab:datasets} summarizes the dataset sizes and realized
partitions. Disease uses the HGCN binary split routine, which
balances the validation and test classes and assigns the remaining
nodes to training; the resulting partition is not a literal 30/10/60
split. Airport uses disjoint per-class splits. Split membership
varies with the seed for Disease and Airport, whereas citation
datasets retain their fixed public partitions. Compared models use
identical splits for each seed, verified by matching split fingerprints.

\begin{table}[htbp]
\centering
\caption{Node-classification datasets and realized partitions.
The ogbn-arxiv experiment uses the official chronological split.
Citation public partitions leave some nodes outside the labeled
train/validation/test sets.}
\label{tab:datasets}
\small
\setlength{\tabcolsep}{5pt}
\begin{tabular}{lrrrl}
\toprule
Dataset & Nodes & Train / validation / test & Classes & Metric \\
\midrule
Disease & 1,044 & 748 / 42 / 254 & 2 & Macro-F1 \\
Airport & 3,188 & 2,232 / 478 / 478 & 4 & Accuracy \\
Cora & 2,708 & 140 / 500 / 1,000 & 7 & Accuracy \\
Citeseer & 3,327 & 120 / 500 / 1,000 & 6 & Accuracy \\
PubMed & 19,717 & 60 / 500 / 1,000 & 3 & Accuracy \\
ogbn-arxiv & 169,343 & 90,941 / 29,799 / 48,603 & 40 & Accuracy \\
\bottomrule
\end{tabular}
\end{table}
\FloatBarrier

\subsection{Disease-NC Data}

The attention-score ablation study (Table~\ref{tab:ablation-score}) uses the official
HGCN Disease-NC graph (\texttt{disease\_md.edges.csv}, 1\,044 nodes, 1\,043 edges)
with the standard 1{,}000-dimensional node features and binary infection-status
labels bundled with Disease-NC in the released HGCN benchmark (same splits as Tables~\ref{tab:datasets} and~\ref{tab:nc}).
The realized partition is 748/42/254, as in Table~\ref{tab:datasets}.

\subsection{LapPE Computation}

We compute the first $k$ non-trivial eigenvectors of the normalised graph
Laplacian $L_{\text{sym}} = I - D^{-1/2}AD^{-1/2}$ using \texttt{scipy.sparse.linalg.eigsh}
(smallest eigenvalues, $k=8$, tolerance $10^{-4}$).
The constant eigenvector (eigenvalue $\approx 0$) is excluded.
Signs are fixed by requiring the column entry with largest absolute value to
be positive.

\subsection{Attention Mask}

The 2-hop neighbourhood mask is computed as a dense $N\times N$ float
tensor: 0.0 for pairs within 2 hops (attend), $-\infty$ outside (blocked).
It is passed as an additive attention mask to both models.

\subsection{Evaluation and Reporting}
\label{app:experimental-reporting}

Node-classification results use validation-selected checkpoints and
report mean $\pm$ population SD over ten seeds (42--51).
The ogbn-arxiv evaluation uses the official chronological split and
reports mean $\pm$ sample SD over five seeds (42--46).
Paired differences are reported in percentage points with nominal
95\% Student-$t$ intervals over matched seeds. These intervals are
not adjusted for multiple comparisons, and repeated runs on one graph
do not quantify uncertainty across graphs. Previously inspected test
sets should not be interpreted as fresh holdouts.
\subsection{Large-Scale Evaluation}

We compare Busemann-modulated linear attention with native
Hypformer-large on ogbn-arxiv over five paired seeds.
Validation-selected checkpoints achieve $72.9120\pm0.1138\%$ test
accuracy with Busemann modulation, versus $72.9457\pm0.1162\%$
for native attention (mean $\pm$ sample SD).
The paired difference is $-0.0337$ pp, with a nominal 95\%
confidence interval of $[-0.2819,+0.2145]$ pp; this establishes
neither an accuracy improvement nor equivalence.
The modification adds 131,074 parameters (979,797 versus 848,723)
and increases full-model inference time from 39.97 to 55.09 ms
and peak allocated inference memory from 4,467.0 to 5,128.5 MiB
in this comparison. Thus, the experiment evaluates the complete
attention modification, rather than isolating geometry at fixed
capacity.

The native Hypformer timing in Table~\ref{tab:linear-performance} uses
39.97 ms from the same profile as the modulated model, rather than
39.84 ms from the earlier profile of those checkpoints.
The dictionary-16 minus focused-positive effect is $-0.007$ pp
$[-0.191,0.177]$; dictionary-64 minus dictionary-16 is $+0.101$ pp
$[-0.196,0.399]$. Neither establishes an accuracy gain or equivalence.

\subsection{Qualitative Analysis Protocol}
\label{app:qualitative}

Figure~\ref{fig:visualization} uses the archived Airport LD and Euclidean
checkpoints selected by validation accuracy. Observation-only inference
reproduces their saved test predictions and train/validation/test metrics;
no retraining is performed. Seed 51 was fixed before inspecting the
Airport embeddings, without an Airport seed search. Both panels display
all 478 test nodes from the same 2,232/478/478 train/validation/test split.

We fit centered PCA separately on training-node preclassifier embeddings,
after the origin log map for LD and directly on Euclidean features.
The first two components retain 80.36\% and 63.54\% of training variance,
respectively. We plot the unscaled PCA coordinates with equal coordinate
scales on the two axes of each panel, using the same label-independent
point order and class colors. The models have independent projection
bases and coordinate units; cloud sizes and projected distances are not
directly comparable. For this run, test accuracy is 89.96\% for LD and
90.38\% for the Euclidean model, so apparent PCA separation alone does not
establish a predictive advantage.

\FloatBarrier

\section{Computational Efficiency}
\label{app:efficiency}

\subsection{Boundary-Dictionary Attention}
\label{exp:attention-scaling}
\label{sec:convergence}

We compare materialized global HSA with materialized and associative
evaluation of the same dictionary kernel. All operators use identical
synthetic Lorentz inputs (seed 42), float32, batch size one, two heads,
and 256 spatial channels per head; the dictionary has 256 directions
per head and one scale. Measurements use an H200 with PyTorch 2.7.1/CUDA
12.6, five warmups and 20 synchronized repetitions, with TF32 disabled.
Timing includes feature construction, aggregation, and Lorentz projection,
but excludes input projections, transfers, residuals, and the GNN.

\begin{table}[htbp]
\centering
\caption{Attention-only inference: median milliseconds / peak
PyTorch-allocated MiB. Dictionary columns evaluate the same finite
kernel; repetitions are timing measurements, not independent seeds.}
\label{tab:attention-scaling-inference}
\label{tab:linear-cost}
\small
\setlength{\tabcolsep}{5pt}
\begin{tabular}{rrrr}
\toprule
$N$ & Dense HSA & Dict.\ materialized & Dict.\ associative \\
\midrule
512 & 0.55 / 41.5 & 1.38 / 43.0 & 1.48 / 42.0 \\
8,192 & 6.26 / 1,384.4 & 5.28 / 632.8 & 1.51 / 177.4 \\
65,536 & 365.03 / 82,531.0 & 229.98 / 33,507.0 & 5.12 / 1,188.0 \\
\bottomrule
\end{tabular}
\end{table}

At 65,536 nodes, associative evaluation is $44.9\times$ faster and
uses $28.2\times$ less peak memory than materializing the same
dictionary kernel. Forward/backward time decreases from 642.56 to
17.22 ms and memory from 51,401.8 to 2,635.1 MiB; dense HSA runs out
of memory. At 512 nodes, association is slightly slower.
The materialized references are not fused or streaming implementations.
These are operator-level savings: the dictionary changes the HSA kernel,
so they do not establish accuracy-preserving linearization or whole-model
speedups.

\subsection{Exact Sparse Execution}

Unlike the dictionary extension, sparse execution preserves masked
HSA in exact arithmetic. On PubMed, the two-hop support contains
1,184,067 pairs (0.3046\% of all pairs). Both paths use the same source,
H200, seed 42, $D=64,H=4,L=2$, no GNN, and 132,382 parameters.

\begin{table}[htbp]
\centering
\caption{Same-source PubMed comparison. Memory and epoch time use one
run; accuracy comes from a separate ten-seed study
(mean $\pm$ population SD).}
\label{tab:sparse-runtime}
\small
\setlength{\tabcolsep}{5pt}
\begin{tabular}{lrrr}
\toprule
Execution & Peak MiB & Seconds/epoch & Test accuracy (\%) \\
\midrule
Dense masked & 64,652.35 & 1.11129 & $76.31\pm0.88$ \\
Sparse supported pairs & 4,013.50 & 0.08355 & $76.37\pm1.01$ \\
\bottomrule
\end{tabular}
\end{table}

Sparse execution reduces peak memory by $16.1\times$ and epoch time
by $13.3\times$. The close accuracy means do not establish predictive
equivalence: reduction order and dropout can alter training trajectories.
These savings concern a fixed graph support, not unrestricted global HSA.

\paragraph{Whole-model costs.}
Table~\ref{tab:linear-performance} reports ogbn-arxiv inference with
GPU-resident inputs, two warmups, and ten synchronized forwards.
Dictionary-16 uses less memory than native Hypformer-large but is
slower in the measured forward pass; operator-level savings therefore
need not translate into end-to-end speedups.

\FloatBarrier
\section{Additional GNN-Branch Controls}
\label{app:arch-ablation}

\paragraph{Component-ablation protocol.}
Table~\ref{tab:gnn-branch} uses seeds 42--51, $D=64,H=4,L=2$, fixed unit attention
curvature, and the Busemann-logit readout. All three arms use Adam with
learning rate $5\times10^{-4}$, weight decay $10^{-4}$, dropout 0.1,
class-weighted cross-entropy, gradient clipping at one, at most 500 epochs,
and patience 80. Checkpoints are selected by validation macro-F1.

\paragraph{Additional controls.}
Separate backbone-only and GNN-only controls further distinguish
whole-branch effects from score replacement. On Disease, the
backbone-only, GNN-only, LD-only, and combined models achieve
67.14\%, 84.76\%, 92.66\%, and 90.85\% macro-F1, respectively.
Adding LD attention to the GNN-only model yields $+6.09$ pp
$[2.81,9.36]$ on Disease and $+2.02$ pp $[1.16,2.88]$ on
PubMed. These additions change capacity and receptive support;
they do not isolate the Busemann score. The recovered branch
controls span source revisions and hardware, and are not used
for cross-device cost comparisons.

\section{Conceptual Comparison}
\label{app:comparison}

The geometric primitive at the heart of BusemannFormer is the use of the
Busemann function as a \emph{dynamic, per-token} self-attention score.
Several prior works also touch the Lorentz model or the Busemann function,
but each of them stops short of this primitive.
Table~\ref{tab:comparison} situates HSA against the four most directly related
methods along two axes: (i) what plays the role of a query and how the
attention/score function is computed, and (ii) which use case the method is
designed for. The reader should take three points from the comparison.
First, Hypformer~\citep{yang2024hypformer} replaces every Euclidean operation
with its Lorentz counterpart, including a hyperbolic dot-product-like score,
but the score remains a pairwise compatibility measure between two embedded
points; no token ever picks a direction on the ideal boundary.
Second, HBNN~\citep{chen2026hbnn} and the Busemann-prototype family
\citep{ghadimi2021busemann,fan2023horosphere} \emph{do} use Busemann functions,
but always with directions $v$ that are fixed (per-class prototypes,
per-layer parameters), so the score is not query-dependent and is not used as
a token-to-token attention kernel.
Third, EmBolic~\citep{kapic2026embolic} uses Busemann-energy scores in an
attention-like way for emotion classification, but the energies are
class-level rather than token-level.
HSA is, to our knowledge, the first construction in which each query token
selects its own ideal-boundary direction $v_i$ from its own embedding and
scores arbitrary key tokens by horospherical depth along $v_i$, making
the Busemann function a genuine self-attention kernel rather than a layer
or a classification head.

\begin{table}[!htbp]
\centering
\caption{Conceptual comparison with closest prior works.}
\label{tab:comparison}
\footnotesize
\setlength{\tabcolsep}{5pt}
\renewcommand{\arraystretch}{1.12}%
\begin{tabularx}{\linewidth}{@{} >{\raggedright\arraybackslash}p{0.20\linewidth} >{\raggedright\arraybackslash}X >{\raggedright\arraybackslash}p{0.19\linewidth} @{}}
\toprule
\textbf{Method}
  & \textbf{Query object and score type}
  & \textbf{Use case / role} \\
\midrule
Hypformer~\citep{yang2024hypformer}
  & Queries and keys embed as Lorentz points; attention uses a hyperbolic
  counterpart of a dot-product score (pair-compatibility within the manifold).
  & Graph Transformer \\[2pt]
HBNN~\citep{chen2026hbnn}
  & Queries are Lorentz points but Busemann terms use \emph{fixed} directions $v$.
  Layers give class logits or FC features, not a dynamic Q-dependent axis.
  & FC \& classification head \\[2pt]
Busemann prototypes~\citep{ghadimi2021busemann}
  & Class prototypes on $\partial\mathbb{L}$ define \emph{Busemann centred}
  losses; no per-token boundary choice from embeddings.
  & Prototype classifier \\[2pt]
EmBolic~\citep{kapic2026embolic}
  & Attention-like scores use coarse \emph{class-level} Busemann energies,
  not per-token dynamic boundary directions.
  & Emotion / fixed labels \\[2pt]
\textbf{Ours (HSA)}
  & For each query $i$: choose $v_i$ \emph{dynamically};
  score $-\tau_h B_{v_i}(k_j)$ is a genuine token-token attention kernel.
  & \textbf{General self-attention} \\
\bottomrule
\end{tabularx}
\end{table}

In short, prior hyperbolic Transformers keep the dot-product abstraction and
prior Busemann-based methods keep the boundary direction fixed; HSA is the
combination that has, to our knowledge, not been considered:
a token-to-token attention score in which each query dynamically selects its
own boundary direction and ranks keys by horospherical depth along it.
This is what enables the flat-limit recovery of scaled dot-product attention
(Proposition~\ref{prop:euclidean-limit}) and the horosphere level-set
characterisation (Appendix~\ref{app:horosphere-levels}) that drive the rest of
the analysis.
\FloatBarrier

\end{document}